# An Intelligent Approach for Negotiating between chains in Supply Chain Management Systems

Shahab firouzi

firouzi@iauyazd.ac.ir

Department of Computer engineering, Yazd Branch, Islamic Azad University, Yazd, Iran

Amin Nezarat

Department of Computer engineering, Yazd Branch, Islamic Azad University, Yazd, Iran

aminnezarat@gmail.com

### ABSTRACT

*Holding commercial negotiations and selecting the best supplier in supply chain management systems are among weaknesses of producers in production process. Therefore, applying intelligent systems may have an effective role in increased speed and improved quality in the selections .This paper introduces a system which tries to trade using multi-agents systems and holding negotiations between any agents. In this system, an intelligent agent is considered for each segment of chains which it tries to send order and receive the response with attendance in negotiation medium and communication with other agents .This paper introduces how to communicate between agents, characteristics of multi-agent and standard registration medium of each agent in the environment. JADE (Java Application Development Environment) was used for implementation and simulation of agents cooperation.*

*Keyword(s):* e-Commerce, e-Business, Supply Chain Management System(SCM), eSCM, Intelligent Agents, JADE, Multi Agents

## INTRODUCTION

A supply chain is a complex and network of facilities and distribution channels which are responsible for provide and distribute materials throughout the chains. Materials can be raw materials, semi-finished product or final product. Primary and upstream chains in supply chain need raw materials and downstream chains send final product to consumer.

Supply chain management applications in production industries and their supply chains consist of very complex techniques. Since there are also many other industries and chains in a supply chain of production industries, material distribution and supply management is one of the basic problems of these industries.

Up to now, they have tried to increase speed in transfer chains messages using new technologies like internet and provide relationship between chains in the least time as possible. In current supply chain management methods (SCM), one central management is often used, that is, one of the chains which is nearer to chain center is recognized as manager of chains and tries to communicate with other chains using supply chain management systems.

Supply chain management tries to solve one of the problems and discussions in production and operation management. Many published contexts mentioned to high supply chain potential in cooperation process between chains and tried to use this potential in providing combination of whole supply chain.

On the basis of definition introduced by Kristopher: "supply chain is the network of organizations that are involved, through upstream and downstream linkages, in the different processes and activities that produce value in the form of products and services in the hands of the ultimate consumer [2]", but modern supply chains move towards converting organizations into chains and see a supply chain as the set of chains which can negotiate each other and move materials between each other.

Therefore, in this paper we have tried to develop a new solution to communicate in chain supply chain using negotiation concept between chains.

## PROBLEM IN COTROLLING CHAINS INTENSIVELY

As a whole, controlling distribution systems using a central control system has several disadvantages such as information security, lack of robustness, scalability, slowness and dependency.

Controller should has extensive knowledge of whole system and chains continually and the very extensive communication from controller to whole chains in different times can slow sending and receiving messages and since there is more dependence of whole supply chain on an intensive system, the flexibility of this complex is reduced and it is hard to repair and compensate damage if there is an event, and a small damage to one of the communicating channels can lead to bigger damage to whole chain and system [3].

Privacy and information security is one of the other concerns about intensive systems which influence into one of the communicating channels can lead to influence into whole system and endangers the security of whole supply chain.

## The existing coordination structure between chains

Usually, material procurement processes in pre-construction stage are well specified to satisfy the owner's and designer's requirements. A general contractor is the main coordinator in these processes focused on materials that will be used in a critical path. In practice, interactions among participating project teams are limited only to closely related teams. This is because it is almost impossible to communicate with dozens of subcontractors, hundreds of suppliers, and thousands of sub-suppliers in a single project.

Since all participants have competing goals and objectives, the lack of coordination and information sharing among them leads to the distorted demand fluctuation known as the "Bullwhip effect"[6]. Such fluctuation could result in excessive inventory, poor product forecasts, insufficient or excessive capacities, long backlogs, uncertain production planning, and high costs for correction such as for expedited shipment and overtime [6]. A practical SCM approach used by some engineering and contracting companies is based on a constant safety lead-time, which is applied to the whole duration of construction phases. This method treats all kinds of projects equally regardless of their different characteristics. In addition, this method does not consider variability involved in delivery and construction processes.

## Intelligent agents and multi-agent systems in supply chain

To address the issue, it seems that multi agent system (MAS) is a good candidate. MAS is system which is composed of several intelligent agents (IA) in interaction with each other [7]. Such systems are used when it cannot be managed via a single IA or it is difficult to control; for instance, online business and designing social structures. In [7], the important characteristics of agents of MAS are listed as:

1. Choice: the agents are at least partially independent.

2. Local view: no agent has a complete view of the system. Usually the system is too complicated for a single agent to make the most of the system from one view point.

3. No centrality: no central agent controls the other agents

In other words, MAS is composed of several independent and synchronic agents which cooperate with each other to reach a common goal. IAS (intelligent agent s) existing in such a system are connected with each other via message communication. The language of these messages is ACL .One of these languages is ebXML as discussed before. The messages that agents send to each other have different meanings such as comparison, commands, or structure. On the other hand, web services are the available software elements for the agents which will give useful information to users. So they are useful for IAS.

Finally, these IASs will be more powerful in a complicated situation dealing with bulk of information to help users. It also can take the responsibility of collecting information, buyers, sellers, and preparation.

There are different rules on the activity of agents which must be observed when something occurs during the time the agents are in touch with one another, for instance, when a suggestion is offered or an answering time ends with no reply. These rules are divided into 3 categories [8]:

**Suggestion rules:** the function of this group is to control and manage the suggestions and requests expressed by other agents .It should be clear if these suggestions are in accordance with the main mechanism of the communication among agents both conceptually and structurally and the requests are analyzed both structurally and conceptually. For example, here it will be decided if the request is allowed to be sent or not.

**Information rules:** in each agent, the receiver of the information is necessarily a receiver of data whose job is to analyze them .These rules controls the condition within which this information is made. Indubitably, this is necessary in the process of communication. The changes in the situation of requests and communication are done within the limits of these rules.

**Clarification rules:** the final goal of each negotiation among the intelligent agents is to complete one or several business among the parties of the negotiation. The last group of rules registers and calculates this business, totally it controls the negotiations.

Since supply chain is a distributed system, using multi-agent systems can be a good approach for supply chain management. In multi-agent systems, any agent is referred to a facility to manage supply chain and the communicating protocols between agents are also defined. Such agents interact with each other to access dynamically, transfer and evaluate supply chain information. For this purpose, architecture (figure 1) should be designed for above mentioned multi-agent system which supply chain management system infrastructure is provided and coordination mechanisms become clear between agents.Researchers who have used these methods in their studies are presented in table 1.

In this section we provide a high-level overview of extant agent applications, with a particular emphasis on a framework to relate them with this present work.It is informative to group extant agent applications into fourclasses: 1) information filtering agents, 2) information retrieval agents, 3) advisory agents, and 4) performative agents. Briefly, most information filtering agents are focused on tasks such as filtering user-input preferences for e-mail [9], network news groups [10], frequently asked questions [11] and arbitrary text [12]. Information retrieval agents address problems associated with collecting information pertaining to commodities such as compact disks [13] and computer equipment , in addition to services such as advertising and insurance. We also include the ubiquitous Web indexing robots in this class along with Web-based agents for report writing, publishing and assisted browsing .

A third class of agents is oriented toward providing intelligent advice. Examples include recommendations for CDs [9], an electronic concierge, an agent "host" for college campus visits [10] and planning support for manufacturing systems.

Agents for strategic planning support , software project coordination and computer interface assistance are also grouped in this class, along with support for military reconnaissance and financial portfolio management [10]. Performative agents in the fourth class are generally oriented toward functions such as business transactions and work performance. Examples include a marketplace for agent-to-agent transactions and an agent system for negotiation , in addition to the performance of knowledge work such as automated scheduling, cooperative learning and automated digital services.

In some production industries, a supply chain should manage hundred products, many suppliers and thousands distributors which is done by a person or group abroad and this process can be well executed by intelligent agents.

In selling processes, an agent can help to increase sale rate and do orders automatically by finding requirements of consumers and receiving orders and managing its financial routines.

In figure 2, a conceptual diagram from agents-based supply chain is presented in which any agent transfers its intended information to other agents (if necessary) by internet or intranet. Every chain in this chain has a specific agent which uses a common dictionary for smoothing the communication of agents. All levels of a chain including raw material producer, storing, transportation, production, distribution and sale use the related agent.

In this model, all information required for agents including stock level, sale data, demand expectation, delivery schedule should be available for agents in different levels.

Any agent tries to collect and build a specific data base using this information by which it can determine the routine of making a product from beginning to the end. Based on such information, an agent shows production schedule and obtains the approximate time of responding to a group and finally can has good flexibility in responding to other chains.

As what said, we introduce a framework to show relation between agents and how do they cooperate with each other. In this framework we have used three aspects: intelligence, collaboration and mobility. To make clearer the framework in figure 3, we present an example.

Assume that many expert systems are working in a very high intelligence level and can continue their work in remote and different spaces. But they aren't able to communicate and cooperate to solve a big and distributed problem. To solve this problem, agents can have cooperation ability between each other in addition to mobility and intelligence. We can communicate with an agent using programming languages like Java and telescript and send orders for him. Any agent can send a demand or order to other agent which is more related to their cooperation than intelligence[5].

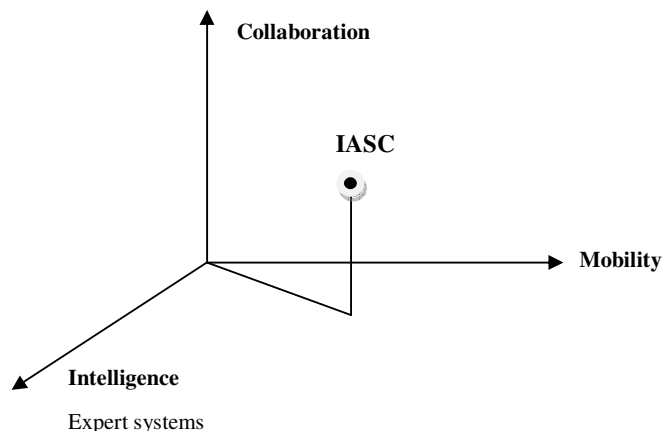

Expert systems

We observe that three agents: intelligence, mobility and collaboration should be simultaneously in different chain levels to manage between chains.

## Conclusion

In this paper, we tried to study and introduce their problems and deficiencies by examining the current methods in designing supply chain and present a new model in communicating between chains in supply chain management systems. In this model which intelligent agents are used for negotiating and making decision in each chain of chain, any agent can access to its intended information throughout the chain and obtain new information and insert information in other chain with movement between chains. In addition, intensive management is omitted in this method and as necessary in different section of production routine, any agents can take responsibility of chain management and this management is continuously moved throughout chain. Therefore, we find that using intelligent agents in supply process is more efficient than traditional supply chain management systems. The authors plan to continue their research about using data mining methods in intelligent agents to improve decision making routine of these agents.

International Journal of Distributed and Parallel Systems (IJDPS) Vol.3, No.3, May 2012

**Figure 1. Architecture**

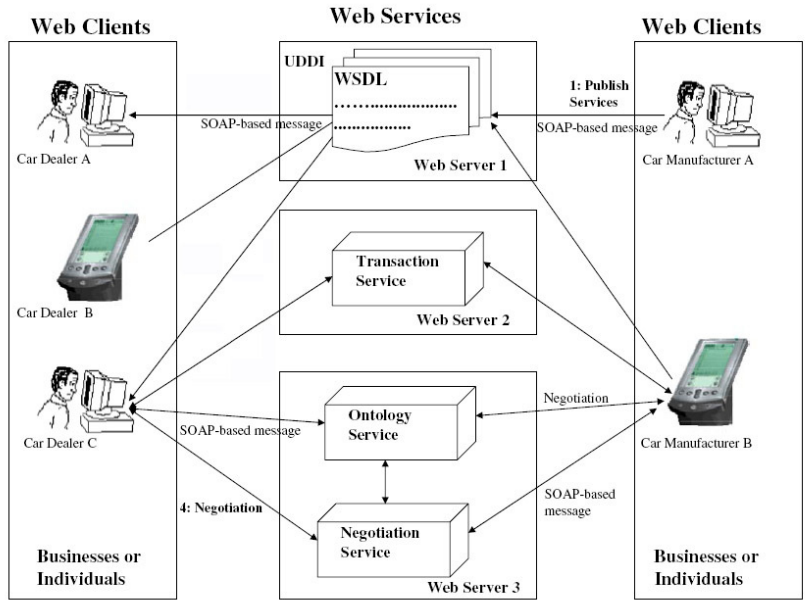

**Table 1.**

| Research domain | Literature |
|---|---|
| Architecture design | Barbuceanu and Fox 1996; Shen et al. 1998;Chen et al. 1999; Fox et al. 2000; Yung et al.2000; Frey et al. 2003; Wagner et al. 2003 |
| Coordination mechanism | Thizy 1991; Lee and Billington 1993; Millar and Yang 1993; Beck and Fox 1994; Ertogral and Wu 2000; Qinghe et al. 2001; Fan et al. 2003; Fink 2004 |
| Supply chain formation | Walsh and Wellman 1999; Walsh et al. 2000; Walsh and Wellman 2003; Babaioff and Walsh 2005 |
| Simulation | Swaminathan et al. 1998; Fu and Piplani 2000; Akkermans 2001; Gjerdrum et al. 2001; Yi et al. 2002; Kumara et al. 2003; Pathak et al. 2003 |

**Figure 2. Conceptual digram of agent-based supply chain automation**

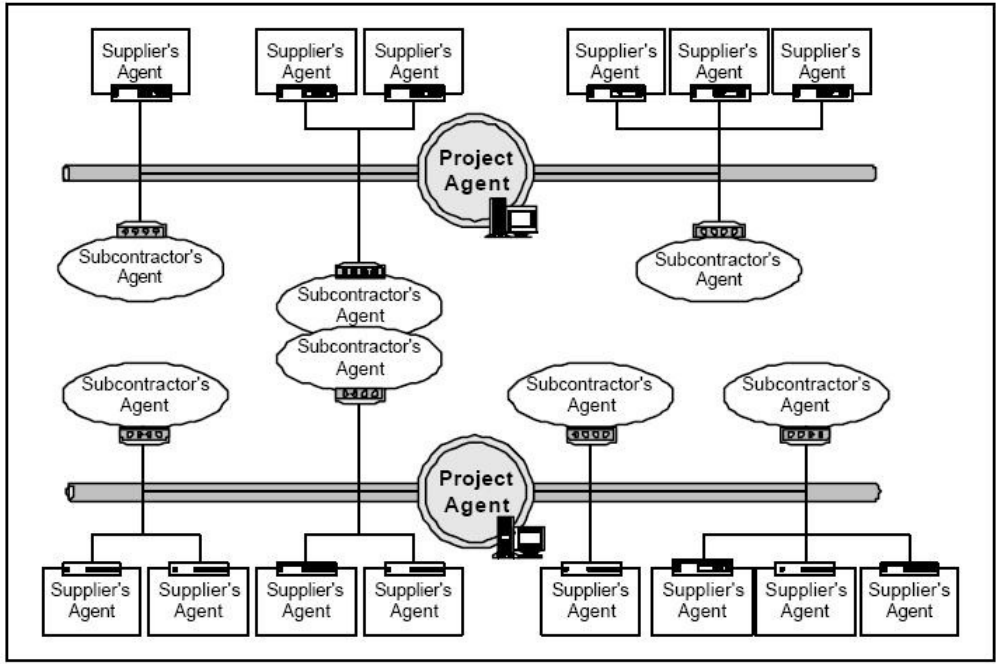

figure1.JPG

**Figure 3. Causal loop diagram**

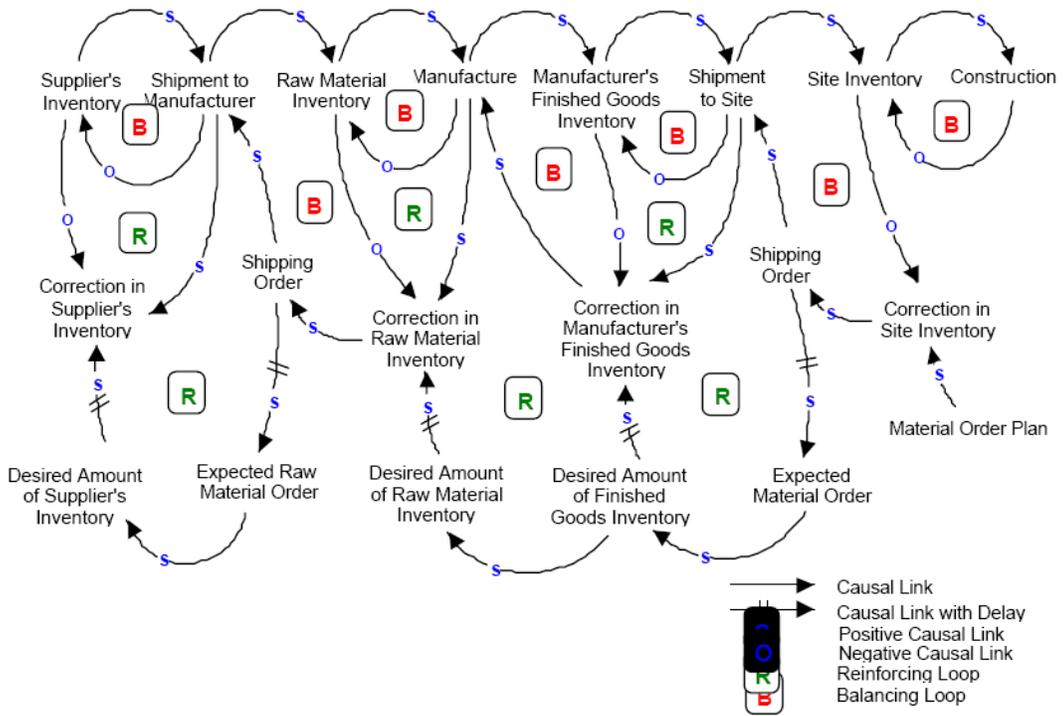

figure2.JPG